\documentclass{article}

\usepackage{paper}
\usepackage[T1]{fontenc}
\usepackage[utf8]{inputenc}
\usepackage{natbib}
\setcitestyle{authoryear,round,citesep={;},aysep={,},yysep={;}}
\usepackage{amsmath,amssymb,amsfonts}
\usepackage{booktabs}
\usepackage{array}
\usepackage{longtable}
\usepackage{graphicx}
\usepackage{xcolor}
\usepackage{microtype}
\usepackage{hyperref}
\usepackage{url}
\usepackage{xspace}
\usepackage{colortbl}
\usepackage{enumitem}
\usepackage{pifont}
\usepackage{tabularx}

\definecolor{markgreen}{HTML}{237A45}
\definecolor{markred}{HTML}{9B2C36}
\colorlet{resourceblue}{paperblue}

\definecolor{taskheader}{HTML}{E5EDF5}
\definecolor{taskgroup}{HTML}{F1F4F8}

\definecolor{resultheader}{HTML}{F1F3F5}
\definecolor{resultours}{HTML}{EAF2FA}

\newcommand{\cmark}{\textcolor{markgreen}{\ding{51}}}
\newcommand{\xmark}{\textcolor{markred}{\ding{55}}}

\newcommand{\benchmark}{RoboCasa-Mem\xspace}

\title{\benchmark: Benchmarking Environment and Process Memory in Long-Horizon Mobile Manipulation}

\author{
    \begin{minipage}[t]{0.96\textwidth}
        \raggedright\normalfont
        Sihang Chen\textsuperscript{1}\quad
        Chen Ye\textsuperscript{2}\quad
        Wei Chen\textsuperscript{1}\quad
        Xiaohan Lei\textsuperscript{1}\\[3pt]
        Zhipeng Tang\textsuperscript{1}\quad
        Sha Zhang\textsuperscript{3}\quad
        Yuan Zhang\textsuperscript{4}\quad
        Zixuan Chen\textsuperscript{5}\\[3pt]
        Jiajun Deng\textsuperscript{1,*}\quad
        Houqiang Li\textsuperscript{1}\quad
        Yanyong Zhang\textsuperscript{1,*}\\[7pt]
        \small
        \textsuperscript{1}\href{https://ustc.edu}{University of Science and Technology of China}\\[2pt]
        \textsuperscript{2}\href{https://tongji.edu.cn}{Tongji University}\quad
        \textsuperscript{3}\href{https://cuhk.edu.hk}{The Chinese University of Hong Kong}\\[2pt]
        \textsuperscript{4}\href{https://ahu.edu.cn}{Anhui University}\quad
        \textsuperscript{5}\href{https://nju.edu.cn}{Nanjing University}\\[4pt]
        \footnotesize\textsuperscript{*}Corresponding authors.
    \end{minipage}
}
\date{}

\hypersetup{
    hidelinks,
    pdftitle={RoboCasa-Mem: Benchmarking Environment and Process Memory in Long-Horizon Mobile Manipulation},
    pdfauthor={Sihang Chen; Chen Ye; Wei Chen; Xiaohan Lei; Zhipeng Tang; Sha Zhang; Yuan Zhang; Zixuan Chen; Jiajun Deng; Houqiang Li; Yanyong Zhang},
    pdfsubject={},
    pdfkeywords={}
}

\begin{document}

\maketitle

\begin{abstract}

In the real world, mobile manipulation robots must perform complex object interactions and make multi-step decisions, often relying on task-relevant information that is no longer observable. However, existing benchmarks provide limited evaluation of such memory-dependent decision-making in mobile manipulation. 
To address this gap, we introduce \textbf{\benchmark}, a new benchmark comprising nine mobile manipulation tasks that require agents to use interaction history at key decision points.
Specifically, the tasks are grouped into two categories of memory: One is environment memory, which captures spatial locations, object attributes, and object relationships. The other is process memory,  which captures single-task progress and multi-task scheduling.
Particularly, we introduce \emph{Behavior-Conditioned Evaluation}. For tasks with multiple valid execution paths, instead of fixing a single target or trajectory, the evaluator derives later task constraints from the robot's earlier actions. This allows us to evaluate whether later behavior is consistent with the robot's own interaction history while reducing shortcuts based on fixed target assignments.
We provide 1,800 successful demonstrations for policy training. We also develop a baseline on our benchmark, namely Hierarchical Memory-Augmented Policy (HMAP). Its high-level planner uses structured entity and process memories to generate subtasks. Its low-level VLA controller uses subgoal-conditioned queries to incorporate VGGT-$\Omega$ geometry into action generation. Together, \benchmark and HMAP baseline support systematic evaluation of memory-dependent behavior and the development of memory-aware mobile manipulation policies.

\end{abstract}

\begin{quote}
    \fontsize{8}{10}\selectfont
    \textbf{Project Page:} \textcolor{resourceblue}{\url{https://sihhangchen.github.io/RoboCasa-Mem/}}\\
    \textbf{Code:} \textcolor{resourceblue}{\url{https://github.com/SiHhangChen/RoboCasa-Mem}}\\
    \textbf{Data:} \textcolor{resourceblue}{\url{https://modelscope.cn/datasets/ouhebai/RoboCasa-Mem/tree/master/data}}
\end{quote}

\section{Introduction}
\label{sec:introduction}

Mobile manipulation requires robots to navigate between locations and interact with objects~\citep{szot2021habitat,nasiriany2024robocasa}. While tabletop manipulation often concentrates task-relevant information within a fixed workspace, mobile manipulation extends interactions across different work areas. As the robot moves and interacts with the environment, previously observed information may become inaccessible yet remain essential for subsequent decisions. For example, when cooking, a robot may need to remember whether it has already added salt to the pot after the salt has dissolved and is no longer visually distinguishable. Such situations motivate maintaining historical information about both the environment and task execution, beyond what is available in the current observation~\citep{torne2026mem}.
Memory dependence is not equivalent to long-horizon manipulation. A task may involve multiple sequential steps without requiring historical information, provided that the current observation is sufficient for decision-making. Conversely, even a short task may require memory when critical information is no longer observable. From a partially observable decision-making perspective, historical information is needed when different interaction histories yield indistinguishable current observations but require different subsequent decisions~\citep{kaelbling1998pomdp}. We therefore evaluate memory based on the necessity of historical information for correct decision-making, rather than task length or execution complexity alone.

Existing mobile-manipulation benchmarks, such as RoboCasa~\citep{nasiriany2024robocasa}, RoboCasa365~\citep{nasiriany2026robocasa365}, MS-HAB~\citep{shukla2025maniskillhab}, BEHAVIOR-1K~\citep{li2024behavior1k}, and the suite of works based on Habitat~\citep{puig2023habitat3, szot2021habitat,habitat19iccv},  primarily evaluate task execution, long-horizon control, generalization, and policy learning, without explicitly isolating decisions that require historical information. In parallel, robotic memory benchmarks such as RMBench~\citep{chen2026rmbench}, RoboMME~\citep{dai2026robomme}, RoboMemArena~\citep{lei2026robomemarena}, and RoboTwin-MeM~\citep{yang2026eventvla} study history-dependent manipulation, predominantly in tabletop or fixed-base settings. 
Mobile manipulation brings these memory demands into settings where robots move between work areas, lose visual access to previously observed entities, and must resume unfinished activities. Two requirements are particularly relevant: \textbf{retaining object states established through earlier interactions but no longer visually distinguishable}, and \textbf{tracking progress and pending actions across interleaved activities in different regions}. Although neither requirement is unique to mobile manipulation, their joint evaluation under changing viewpoints and cross-region execution remains limited.

To systematically characterize these requirements, we organize memory into two complementary dimensions: \emph{Environment Memory} and \emph{Process Memory}.
Environment Memory captures information about the physical environment that cannot be reliably recovered from the current observation, encompassing \emph{Spatial Memory}, \emph{Attribute Memory}, and \emph{Relational Memory}.
Process Memory tracks task execution states and information needed to determine subsequent actions, encompassing \emph{Single-Task Progress} and \emph{Multi-Task Scheduling}.
This taxonomy explicitly accounts for visually unobservable object attribute changes and the scheduling of interleaved tasks, providing a systematic characterization of memory requirements in mobile manipulation.

Based on this taxonomy, we introduce \textbf{\benchmark}, a benchmark for memory in mobile manipulation. 
\benchmark contains nine mobile-manipulation tasks covering the five memory categories above. 
Each task is designed to introduce \emph{observation-insufficient decision points}, where the current observation alone cannot determine the appropriate action, requiring the robot to retrieve and utilize information from its interaction history.
This design explicitly evaluates memory dependence rather than relying on task horizon as its proxy.
For tasks with multiple valid execution paths, we introduce \emph{Behavior-Conditioned Evaluation}, which derives later constraints from the policy's own earlier interactions rather than fixing a target in advance. This allows history-consistent behavior to be evaluated without prescribing a single execution trajectory.

To provide a baseline, we develop a \emph{Hierarchical Memory-Augmented Policy (HMAP)} that combines structured memory for planning with geometry-augmented control. A high-level planner maintains \emph{Entity Memory} and \emph{Process Memory} to track environment states, execution progress, and pending actions, and generates executable subgoals. A low-level VLA controller executes these subgoals using current observations and multi-view geometric features extracted by VGGT-$\Omega$~\citep{wang2026vggtomega}, which are incorporated through task-conditioned queries and action modulation. Across the nine tasks, HMAP improves average success from 16.2\% to 24.2\% over $\pi_{0.5}$~\citep{black2025pi05}, with gains concentrated in hidden-attribute recall. Improvements on process-memory tasks remain inconsistent, highlighting challenges in progress tracking and interleaved execution.

\begin{table*}[t]
\caption{Comparison of simulated robotic memory benchmarks.
Coverage denotes at least one qualifying task, not uniform requirements
across tasks or equivalent taxonomies.}
\label{tab:benchmark-comparison}
\centering
\scriptsize
\setlength{\tabcolsep}{2pt}
\renewcommand{\arraystretch}{1.0}
\setlength{\aboverulesep}{1.5pt}
\setlength{\belowrulesep}{1.5pt}

\begin{tabular*}{\linewidth}{@{\extracolsep{\fill}}l*{8}{c}@{}}
\toprule
& & & \multicolumn{5}{c}{Memory coverage} & \\
\cmidrule(lr){4-8}
Benchmark
& Mobile
& \shortstack{History\\[-1pt]dependence}
& Spatial
& Attribute
& Relation
& Progress
& Scheduling
& \shortstack{Stage\\[-1pt]eval} \\
\midrule

\shortstack[l]{MIKASA-Robo\\[-2pt]
{\scriptsize\citep{cherepanov2025mikasa}}}
& \cellcolor{gray!10}\xmark
& \cmark & \cmark & \xmark & \cmark & \cmark & \xmark & \xmark \\

\shortstack[l]{RMBench\\[-2pt]
{\scriptsize\citep{chen2026rmbench}}}
& \cellcolor{gray!10}\xmark
& \cmark & \cmark & \xmark & \cmark & \cmark & \xmark & \xmark \\

\shortstack[l]{RoboMME\\[-2pt]
{\scriptsize\citep{dai2026robomme}}}
& \cellcolor{gray!10}\xmark
& \cmark & \cmark & \xmark & \cmark & \cmark & \xmark & \xmark \\

\shortstack[l]{RoboTwin-Mem\\[-2pt]
{\scriptsize\citep{yang2026eventvla}}}
& \cellcolor{gray!10}\xmark
& \cmark & \cmark & \xmark & \cmark & \cmark & \xmark & \xmark \\

\shortstack[l]{RoboMemArena\\[-2pt]
{\scriptsize\citep{lei2026robomemarena}}}
& \cellcolor{gray!10}\xmark
& \cmark & \cmark & \cmark & \cmark & \cmark & \xmark & \cmark \\

\midrule
\textbf{\benchmark{} (Ours)}
& \cellcolor{gray!10}\cmark
& \cmark & \cmark & \cmark & \cmark & \cmark & \cmark & \cmark \\
\bottomrule
\end{tabular*}
\end{table*}

Our contributions are threefold:
\begin{itemize}[leftmargin=1.2em]
    \item \textbf{Benchmark and dataset.} We introduce \benchmark, comprising nine mobile-manipulation tasks organized into five memory categories across environment-state and task-state memory, together with 1,800 successful demonstration trajectories for policy training.
    \item \textbf{Evaluation protocol.} We introduce Behavior-Conditioned Evaluation, which derives subsequent semantic constraints from the robot's actual interactions to assess history-consistent behavior across multiple valid execution paths. We report both final task success and stage-level completion.
    \item \textbf{Baseline and findings.} We develop \emph{Hierarchical Memory-Augmented Policy (HMAP)} that combines structured semantic memory for planning with 3D geometric feature for control. Our experiments reveal gains concentrated in hidden-attribute recall and identify persistent challenges in task-progress tracking and interleaved multi-task execution.
\end{itemize}

\section{Related Work}
\label{sec:related-work}

\paragraph{Mobile-manipulation benchmarks.}
Existing benchmarks study diverse aspects of mobile manipulation, including task diversity, long-horizon execution, generalization, and scalable policy learning. RoboCasa and RoboCasa365 provide large-scale household environments, tasks, and demonstrations~\citep{nasiriany2024robocasa,nasiriany2026robocasa365}; Habitat 2.0 and MS-HAB focus on long-horizon household rearrangement and low-level mobile-manipulation control~\citep{szot2021habitat,shukla2025maniskillhab}; and BEHAVIOR-1K covers a broad range of household activities and physical interactions~\citep{li2024behavior1k}. These benchmarks do not explicitly isolate memory dependence as an evaluation target. In contrast, \benchmark focuses on decisions for which the current observation is insufficient and earlier information must be retained.

\paragraph{Memory benchmarks for manipulation.}
Recent benchmarks explicitly evaluate memory in robotic manipulation. MIKASA-Robo and RMBench study memory-intensive manipulation tasks~\citep{cherepanov2025mikasa,chen2026rmbench}; RoboMME evaluates temporal, spatial, object, and procedural memory~\citep{dai2026robomme}; RoboMemArena studies long-horizon manipulation with memory-related annotations~\citep{lei2026robomemarena}; and RoboTwin-MeM considers non-Markovian tasks in which later actions depend on earlier visual observations~\citep{yang2026eventvla}. Most of these benchmarks focus on tabletop or fixed-base manipulation. \benchmark instead targets mobile manipulation and emphasizes memory demands arising from changing environment states and interleaved task execution. It also evaluates history-dependent decisions without requiring a single predefined execution path.

\paragraph{Memory-augmented embodied policies.}
Memory-augmented policies differ mainly in how historical information is selected and represented. MemER retrieves historical keyframes for high-level subtask generation, while EventVLA, KEMO, and UniMem retain task-relevant or event-triggered observations for later prediction and control~\citep{sridhar2025memer,yang2026eventvla,zeng2026kemo,osterberg2026unimem}. Other methods combine memory across temporal or semantic scales: MemoryVLA uses working and perceptual-cognitive memory, MEM combines short-term video memory with long-term textual memory, and Mimir separates world memory from task memory~\citep{shi2025memoryvla,torne2026mem,xu2026mimir}. Our method similarly uses memory at high levels and combines persistent structured memory for high-level task decisions with current 3D visual features for low-level control.

\section{The \benchmark Benchmark}
\label{sec:benchmark}

\begin{figure*}[t!]
    \centering
    \includegraphics[width=\textwidth]{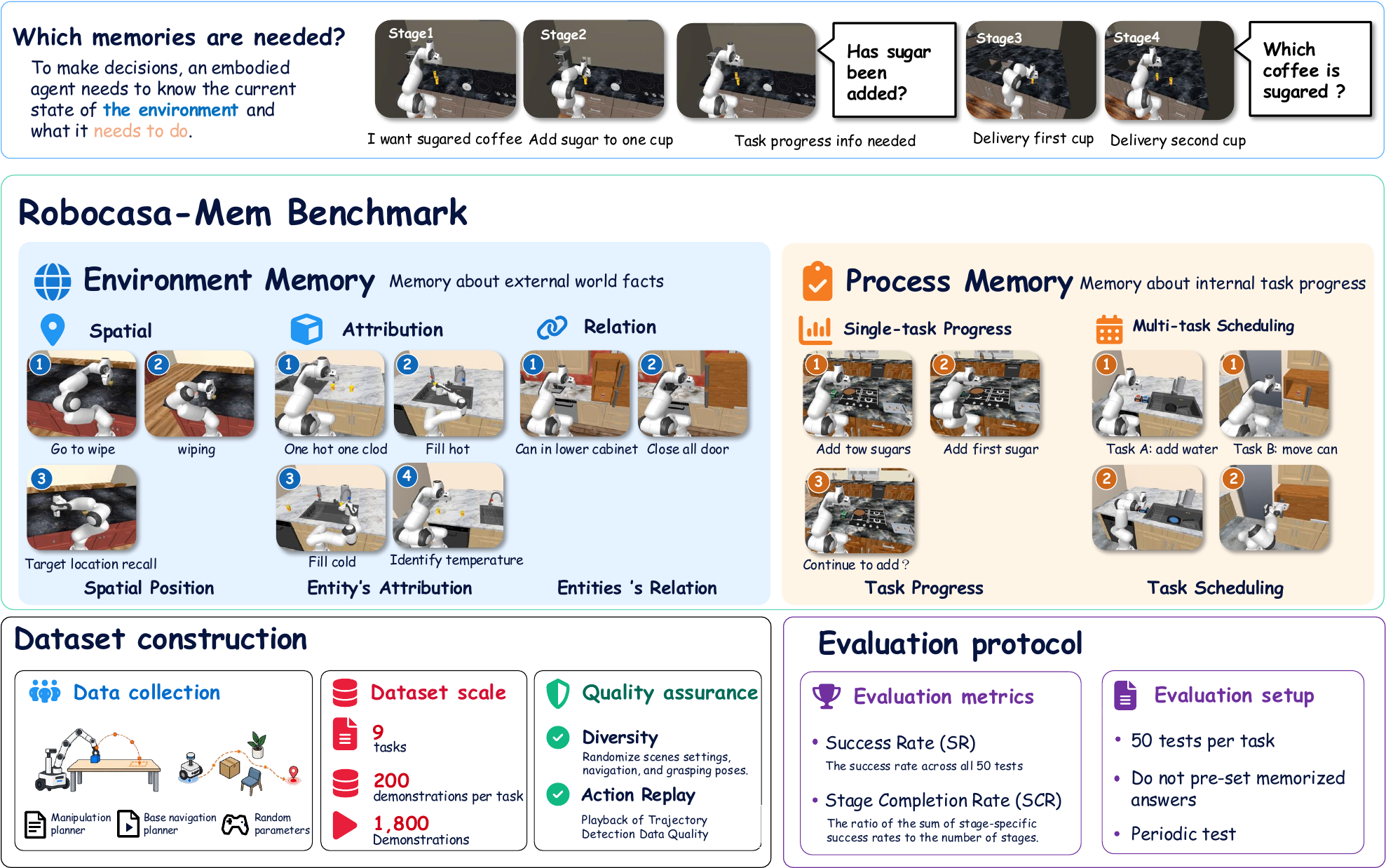}
    \caption{Overview of \benchmark. Top: a history-dependent cup-selection
    example illustrates why the current observation alone may not determine
    the correct action. Middle: five memory categories cover spatial,
    attribute, and relational Environment Memory, together with single-task
    progress and multi-task scheduling in Process Memory. Bottom:
    demonstration construction and semantic evaluation support analysis of
    final outcomes and stage-level completion. Table~\ref{tab:tasks} lists the
    nine core tasks.}
    \label{fig:benchmark-overview}
\end{figure*}

\benchmark contains nine household mobile-manipulation tasks covering five memory types under two groups: \emph{Environment Memory} and \emph{Process Memory}. Figure~\ref{fig:benchmark-overview} gives an overview of the benchmark.

\subsection{Memory Demand and Task Design}
\label{sec:formulation}

\paragraph{Memory-dependent task design.}
We design tasks such that information required for a later decision is observed earlier but cannot be reliably recovered from the current observation. For example, after sugar dissolves in water, the current image does not reveal whether sugar was previously added to a cup. The policy must therefore retain information from earlier interactions rather than relying only on the current observation. This distinguishes memory dependence from task horizon: a task may contain many steps without requiring memory if each decision can still be made from the current observation.

We formalize this design principle using observation aliasing in partially observable environments~\citep{kaelbling1998pomdp}. Let $x$ denote the task instruction, $o_t$ the current observation, and $h_t=(o_1,a_1,\ldots,a_{t-1},o_t)$ the interaction history. For a semantic query $q$, let $\mathcal{G}^{\star}_q(h_t,x)$ denote the set of acceptable responses. Historical information is required when two feasible histories satisfy
\begin{equation}
    o_t(h_t)=o_t(h'_t),
    \qquad
    \mathcal{G}^{\star}_q(h_t,x)\cap
    \mathcal{G}^{\star}_q(h'_t,x)=\varnothing.
    \label{eq:memory-demand}
\end{equation}
That is, similar current observations may require different decisions because of different interaction histories. We use this condition as an idealized task-design criterion rather than requiring observations to be exactly identical at the pixel level.

\subsection{Task Classification}
\label{sec:tasks}

The nine tasks are grouped according to the historical information required for later decisions. \emph{Environment Memory} concerns information about the environment that may no longer be recoverable from the current observation. It includes \emph{Spatial Memory}, which retains earlier object locations; \emph{Attribute Memory}, which retains object properties or states; and \emph{Relational Memory}, which retains associations between entities.

\emph{Process Memory} concerns information about task execution. It includes \emph{Single-Task Progress}, which tracks completed actions, search progress, or counts within one task, and \emph{Multi-Task Scheduling}, which tracks pending actions and task progress across interleaved activities. Each task is assigned to its primary memory category, although successful execution may involve more than one type of memory.
Table~\ref{tab:tasks} summarizes the nine core tasks and their main memory requirements. Detailed task definitions, stage predicates, and success conditions are provided in Appendix~\ref{app:tasks}.

\begin{table*}[t]
\caption{Overview of the nine \benchmark tasks, organized into five
memory categories under Environment Memory and Process Memory.
Quoted descriptions summarize the tasks rather than reproduce
the exact policy instructions.}
\label{tab:tasks}
\centering
\footnotesize
\setlength{\tabcolsep}{4pt}
\renewcommand{\arraystretch}{1.05}

\begin{tabularx}{\linewidth}{
    >{\raggedright\arraybackslash}p{0.18\linewidth}
    >{\raggedright\arraybackslash}p{0.15\linewidth}
    >{\raggedright\arraybackslash}p{0.17\linewidth}
    >{\itshape\raggedright\arraybackslash}X
}
\toprule
\rowcolor{taskheader}
\textbf{Task}
& \textbf{Memory type}
& \textbf{Key challenge}
& \textbf{\upshape Description} \\
\midrule

\rowcolor{taskgroup}
\multicolumn{4}{c}{\textbf{Environment Memory}} \\

\textbf{PutBackCup}
& Spatial
& Location recall
& ``Place the cup at the previously cleaned spot.'' \\
\addlinespace[1pt]

\textbf{SweetCup}
& Attribute
& Hidden attribute
& ``Prepare one sweetened and one unsweetened coffee, then select
the sweetened cup.'' \\
\addlinespace[1pt]

\textbf{TempReport}
& Attribute
& Attribute recall
& ``Recall whether hot or cold water was dispensed, then press
the matching button.'' \\
\addlinespace[1pt]

\textbf{CanRecall}
& Relational
& Object--container binding
& ``After closing the cabinets, recall the requested items'
locations and place them back on the tabletop.'' \\
\addlinespace[1pt]

\textbf{MilkMatch}
& Relational
& Source--target binding
& ``Transport milk boxes in two trips, placing each on the plate
matching its source tray's color.'' \\

\midrule
\rowcolor{taskgroup}
\multicolumn{4}{c}{\textbf{Process Memory}} \\

\textbf{DrawerSearch}
& Single-task progress
& Search progress
& ``Count the drawers checked while searching for the red cube,
then press the answer button that many times.'' \\
\addlinespace[1pt]

\textbf{SugarCount}
& Single-task progress
& Action counting
& ``Add $K$ sugar cubes to the hot chocolate, then press
the completion button.'' \\
\addlinespace[1pt]

\textbf{HeatAndInspect}
& Multi-task scheduling
& Pending obligation
& ``Inspect drawers while heating a pan, then return to turn
off the stove and submit.'' \\
\addlinespace[1pt]

\textbf{FillAndRetrieve}
& Multi-task scheduling
& Pending obligation
& ``Retrieve a can while dispensing water, then return to turn
off the faucet and complete both tasks.'' \\

\bottomrule
\end{tabularx}
\end{table*}

\subsection{Benchmark Construction}
\label{sec:data}

\paragraph{Environment and policy interface.}
We implement all tasks in RoboCasa, combining mobile navigation with object and fixture manipulation~\citep{nasiriany2024robocasa}. Each episode is associated with a language instruction. At each recorded timestep, the dataset contains three synchronized $224\times224$ RGB views, a 37-dimensional robot state, and a 13-dimensional action at 20 Hz. The three views are captured by the left external camera, right external camera, and wrist-mounted camera. The standard policy receives the language instruction, RGB observations, and robot state, and predicts the robot action.

\paragraph{Demonstration collection.}
We collect demonstrations using a rule-based expert that combines MPLib for arm motion planning with a closed-loop controller for mobile-base navigation. Each task is decomposed into navigation and manipulation stages, and the corresponding controller is used to execute each stage. To increase data diversity, we randomize task-relevant scene configurations, including object positions and appearance, manipulation order, navigation targets, and manipulation poses where applicable. Some task parameters are also varied across episodes, such as the requested number of sugar cubes in SugarCount.

We collect 200 successful demonstrations for each task, resulting in 1,800 trajectories in total. Only trajectories that satisfy the required stage sequence and final semantic predicates are retained. Detailed state and action definitions, controller settings, randomization procedures, trajectory statistics, validation procedures, and dataset format are provided in Appendix~\ref{app:data}.

\subsection{Evaluation Protocol}
\label{sec:evaluation}

\paragraph{Behavior-conditioned evaluation.}
We evaluate each task over 50 held-out scene seeds using the same randomized environments for all policies. For tasks with multiple valid execution paths, the evaluator derives later constraints from the policy's earlier interactions rather than assigning a fixed target in advance. For example, in SweetCup, either cup may receive sugar, and the evaluator later checks whether the policy selects the same cup. Success is defined by task semantics rather than by reproducing the scripted expert trajectory. Detailed stage and terminal predicates are in Appendix~\ref{app:tasks}.

\paragraph{Evaluation metrics.}
We report two complementary metrics. Task Success Rate (SR) measures the fraction of episodes in which the complete task is successfully completed. Stage Completion Rate (SCR) summarizes task progress as the average success rate over the $K$ predefined evaluation stages, i.e., $\mathrm{SCR}=\frac{1}{K}\sum_{k=1}^{K} r_k$, where $r_k$ denotes the success rate of the $k$-th evaluation stage. SR captures final task completion, while SCR provides a stage-level measure of how far a policy progresses through the task.

\section{Hierarchical Memory-Augmented Policy}
\label{sec:method}

\begin{figure}[t!]
    \centering
    \includegraphics[width=0.95\linewidth]{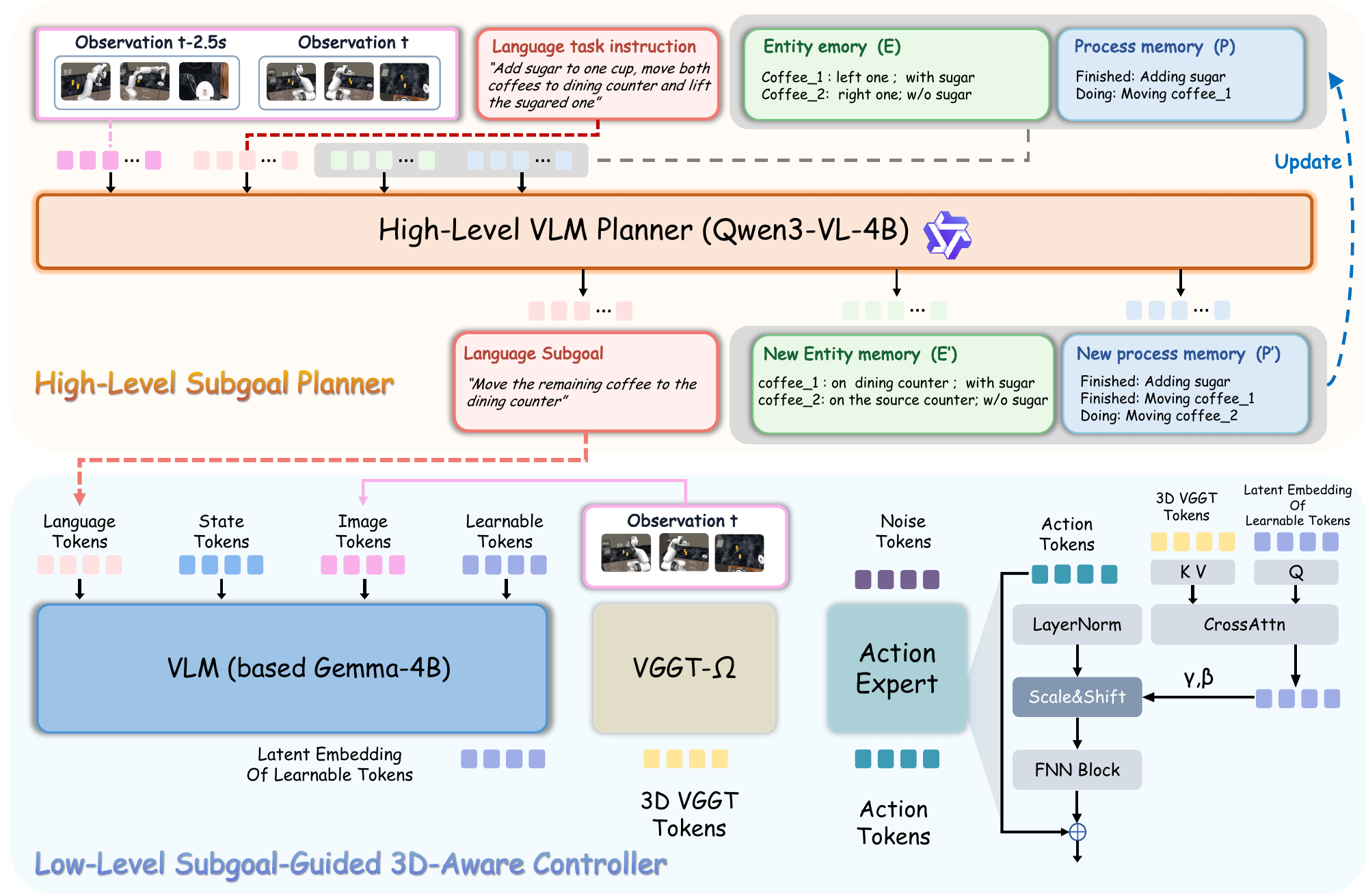}
    \vspace{-0.3cm}
    \caption{\textbf{Overview of our Hierarchical Memory-Augmented Policy.} The high-level planner maintains entity and process memory to select an active subgoal. The low-level subgoal-guided 3D-aware controller conditions on this subgoal and retrieves task-relevant features from current multi-view geometry to modulate action generation.}
    \label{fig:policy-overview}
\end{figure}

Our policy consists of a high-level symbolic memory planner and a low-level subgoal-guided 3D-aware controller. The planner maintains entity memory and process memory to reason over long-horizon task progress and select an executable subgoal. Conditioned on this subgoal, the controller retrieves task-relevant information from current multi-view geometry and uses it to guide action generation. This separation allows symbolic memory to support long-horizon reasoning while geometric context is used for low-level execution.
At each planning time $t_k$, the planner updates entity memory $e_k$ and process memory $p_k$ and produces the current subgoal $g_k$, which is passed to the low-level controller. Figure~\ref{fig:policy-overview} illustrates the overall architecture.

\subsection{High-Level Symbolic Memory Planner}
\label{sec:high-level-memory}

Building on language memory for high-level planning~\citep{torne2026mem}, we maintain two complementary forms of symbolic memory. Entity memory $e_k$ stores task-relevant properties and spatial information about entities, while process memory $p_k$ tracks execution progress and pending steps. Entity memory is intentionally sparse, retaining only information relevant to future task decisions.

We use Qwen3-VL-4B-Instruct~\citep{bai2025qwen3vl} as the planner. At planning step $k$, it takes the overall instruction $x$, previous memories $(e_{k-1},p_{k-1})$, and observations from the current and an earlier timestep:
\begin{equation}
    (e_k,p_k,g_k)
    =H_{\theta}\bigl(
    c,x,e_{k-1},p_{k-1},
    \mathcal{I}_{t_k},\mathcal{I}_{t_k-50}
    \bigr),
    \label{eq:entity-memory-update}
\end{equation}
where $c$ denotes the fixed planner prompt. The two observations provide evidence of recent state changes. The updated memories are recursively passed to subsequent planner calls, while only the executable subgoal $g_k$ is provided to the low-level controller. This recurrent memory allows the planner to preserve task-relevant information that may no longer be observable from the current scene.

The planner is supervised offline using structured annotations derived from demonstration trajectories. The memory schema, annotation procedure, planner prompts, and additional examples are provided in Appendix~\ref{app:method}.

\subsection{Low-Level subgoal-guided 3D-aware  Controller}
\label{sec:low-level-controller}

The low-level subgoal-guided 3D-aware controller builds on $\pi_{0.5}$~\citep{black2025pi05} and augments it with geometric features from a frozen VGGT-$\Omega$ encoder~\citep{wang2026vggtomega}. VGGT-$\Omega$ jointly encodes the current RGB observations from the left external, right external, and eye-in-hand cameras, producing multi-view geometric tokens that remain separate from the standard PaliGemma visual--language--state representation.

To incorporate geometry selectively, we introduce task-conditioned queries into the PaliGemma stream. Conditioned on the current observations, robot state, overall instruction, and active subgoal, these queries retrieve relevant information from the VGGT-$\Omega$ representation through cross-attention. The retrieved geometric features are then used to modulate the corresponding action-expert layers through learned scale and shift transformations. Thus, geometry influences action generation through task-conditioned retrieval and layerwise modulation rather than direct token concatenation.

The planner and controller are trained separately: the planner learns symbolic subgoal prediction and memory updates, while the controller is optimized with the standard flow-matching objective for action generation. At inference, the planner provides the active subgoal, and the controller combines it with current multi-view geometry to execute the task. Detailed architecture, optimization settings, and inference procedures are provided in Appendix~\ref{app:method}.

\section{Experiments}
\label{sec:experiments}

We evaluate our method on the nine core \benchmark tasks, analyze its performance across different memory requirements, validate it on a physical AgileX Cobot Magic Platform, and conduct an ablation study to examine the contributions of the main policy components.

\subsection{Experimental Setup}
\label{sec:experimental-setup}

\paragraph{Baselines and training.}
We compare ACT, Diffusion Policy (DP), $\pi_{0.5}$, MemER-style, and our hierarchical memory policy~\citep{zhao2023act,chi2023diffusion,black2025pi05,sridhar2025memer}. All methods are trained on the same task demonstrations. For a controlled comparison, MemER-style uses the same Qwen3-VL-4B-Instruct high-level model and low-level policy backbone as our method, while differing in high-level memory training. Detailed training schedules, sampling strategies, and per-method hyperparameters are provided in Appendix~\ref{app:method}.

\subsection{Main results}
\label{sec:main-results}

\begin{table*}[t]
\caption{Results on the nine \benchmark tasks.
SR denotes task success rate, and SCR denotes stage completion rate.
Avg.\ is the macro-average across all nine tasks.
Best nonzero values are bolded, including ties at the reported precision.}
\label{tab:main-results}
\centering
\footnotesize
\setlength{\tabcolsep}{3pt}
\renewcommand{\arraystretch}{1.15}
\setlength{\aboverulesep}{2pt}
\setlength{\belowrulesep}{2pt}

\begin{tabular*}{\linewidth}{@{\extracolsep{\fill}}l*{10}{c}@{}}
\toprule
\multicolumn{11}{@{}l}{\textbf{(a) Environment Memory}} \\
\addlinespace[3pt]

\textbf{Model}
& \multicolumn{2}{c}{\textbf{PutBackCup}}
& \multicolumn{2}{c}{\textbf{SweetCup}}
& \multicolumn{2}{c}{\textbf{TempReport}}
& \multicolumn{2}{c}{\textbf{CanRecall}}
& \multicolumn{2}{c}{\textbf{MilkMatch}} \\
\cmidrule(lr){2-3}
\cmidrule(lr){4-5}
\cmidrule(lr){6-7}
\cmidrule(lr){8-9}
\cmidrule(lr){10-11}

\rowcolor{resultheader}
\cellcolor{white}
& SR & SCR
& SR & SCR
& SR & SCR
& SR & SCR
& SR & SCR \\
\midrule

ACT
& 0.00 & 0.48
& 0.00 & 0.07
& 0.00 & 0.09
& 0.00 & 0.09
& 0.00 & 0.03 \\

DP
& 0.00 & 0.20
& 0.00 & 0.04
& 0.08 & 0.21
& 0.00 & 0.05
& 0.00 & 0.05 \\

$\pi_{0.5}$
& 0.68 & \textbf{0.85}
& 0.04 & 0.27
& 0.22 & 0.57
& 0.00 & 0.24
& 0.12 & \textbf{0.47} \\

MemER-style
& 0.10 & 0.22
& 0.00 & 0.21
& 0.00 & 0.58
& 0.00 & 0.21
& 0.00 & 0.19 \\

\midrule
\rowcolor{resultours}
\textbf{HMAP (Ours)}
& \textbf{0.70} & 0.79
& \textbf{0.10} & \textbf{0.46}
& \textbf{0.86} & \textbf{0.93}
& 0.00 & \textbf{0.26}
& \textbf{0.16} & \textbf{0.47} \\
\bottomrule
\end{tabular*}

\vspace{6pt}

\begin{tabular*}{\linewidth}{@{\extracolsep{\fill}}l*{10}{c}@{}}
\toprule
\multicolumn{11}{@{}l}{\textbf{(b) Process Memory and Overall Average}} \\
\addlinespace[3pt]

\textbf{Model}
& \multicolumn{2}{c}{\textbf{DrawerSearch}}
& \multicolumn{2}{c}{\textbf{SugarCount}}
& \multicolumn{2}{c}{\shortstack{\textbf{HeatAnd}\\\textbf{Inspect}}}
& \multicolumn{2}{c}{\shortstack{\textbf{FillAnd}\\\textbf{Retrieve}}}
& \multicolumn{2}{c}{\textbf{Avg.}} \\
\cmidrule(lr){2-3}
\cmidrule(lr){4-5}
\cmidrule(lr){6-7}
\cmidrule(lr){8-9}
\cmidrule(lr){10-11}

\rowcolor{resultheader}
\cellcolor{white}
& SR & SCR
& SR & SCR
& SR & SCR
& SR & SCR
& SR & SCR \\
\midrule

ACT
& 0.00 & 0.08
& 0.04 & 0.07
& 0.00 & 0.25
& 0.00 & 0.26
& 0.00 & 0.16 \\

DP
& 0.00 & 0.08
& 0.14 & 0.19
& 0.00 & 0.23
& 0.00 & 0.18
& 0.02 & 0.14 \\

$\pi_{0.5}$
& \textbf{0.12} & \textbf{0.32}
& \textbf{0.28} & \textbf{0.23}
& 0.00 & 0.32
& 0.00 & \textbf{0.39}
& 0.16 & 0.41 \\

MemER-style
& 0.00 & 0.25
& 0.08 & 0.10
& 0.00 & 0.32
& 0.00 & 0.20
& 0.02 & 0.25 \\

\midrule
\rowcolor{resultours}
\textbf{HMAP (Ours)}
& 0.08 & 0.31
& \textbf{0.28} & 0.20
& 0.00 & \textbf{0.50}
& 0.00 & 0.34
& \textbf{0.24} & \textbf{0.47} \\
\bottomrule
\end{tabular*}
\end{table*}

\subsubsection{Simulation Results}

\paragraph{Overall performance.}
Table~\ref{tab:main-results} summarizes performance on the nine core tasks. Our method achieves the highest average SR and SCR, reaching 0.24 and 0.47 compared with 0.16 and 0.41 for $\pi_{0.5}$. ACT, DP, and MemER-style rarely complete the full tasks despite making partial progress on several stages. In particular, inspection of MemER-style failures shows that many occur around subtask transitions, suggesting that successful execution requires not only manipulation ability but also reliable maintenance and use of task-relevant history.

\paragraph{Environment memory.}
The clearest gains appear when the information needed for a later decision is no longer recoverable from the current observation. On TempReport, our method improves SR from 0.22 to 0.86 and SCR from 0.57 to 0.93 relative to $\pi_{0.5}$. Among episodes reaching the reporting stage, the conditional success rate of recalling the correct temperature increases from 0.41 to 0.90. A similar improvement is observed on SweetCup, where the corresponding conditional recall success increases from 0.22 to 0.45.

The gains are smaller on PutBackCup and MilkMatch at the task level, but stage-level analysis again shows improved history-dependent decisions. For example, the conditional success of returning the cup to its original location increases from 0.72 to 0.83, while the average conditional success of the two color-matched milk placements increases from 0.47 to 0.79. These results indicate that explicit memory primarily improves decisions that require retrieving earlier environment states, rather than uniformly improving all manipulation stages. Stage-wise results are detailed in the Appendix.

\paragraph{Process memory.}
A different pattern emerges for Process Memory. On Single-Task Progress tasks, our method and $\pi_{0.5}$ perform similarly, suggesting that explicit memory provides limited benefit when progress remains partially inferable from the ongoing execution state. Multi-Task Scheduling is substantially harder: on HeatAndInspect, our method improves SCR from 0.32 to 0.50, but no method completes the full task. FillAndRetrieve remains similarly challenging. These results suggest that multi-task execution requires not only retaining unfinished obligations, but also correctly switching between and resuming task threads.

\subsubsection{Real-Robot Evaluation}
\label{sec:real-robot-evaluation}

We further evaluate the policies on the AgileX Cobot Magic platform using an Object Transport task. The robot picks an object from a colored source container, turns toward the opposite workspace, and places it into a target container matching the source container's color. Once the robot turns, the source container leaves the camera view, requiring the policy to retain its color for subsequent target selection. We collect 100 demonstrations and train ACT, DP, $\pi_{0.5}$, and HMAP using the same policy interface as in simulation. As shown in Table~\ref{tab:cobot-SingleObjMatch}, HMAP completes 8 of 10 trials, compared with 2 of 10 for $\pi_{0.5}$, with the difference concentrated at the final target-selection stage.

\begin{figure}[t]
    \centering
    \includegraphics[width=\linewidth]{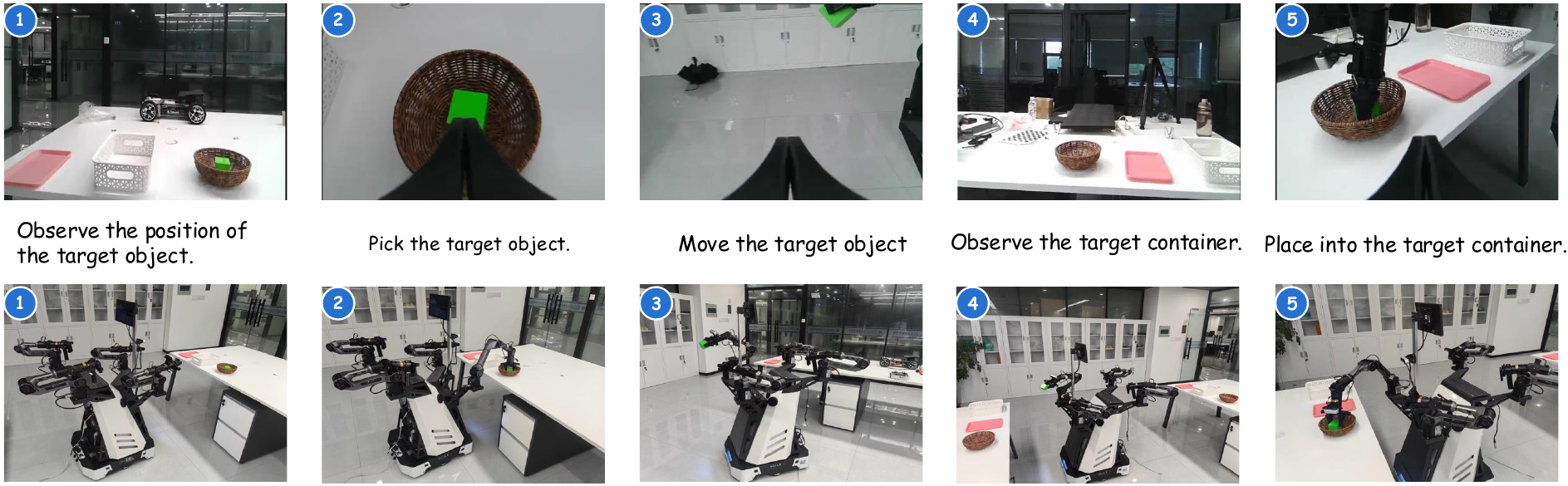}
    \vspace{-0.5cm}
    \caption{An illustration of our real-world experiments conducted on AgileX Cobot Magic Platform. The task is set as: the robot first observes and picks up an object, then turns to transport it and places it into the corresponding container.}
    \label{fig:real-task}
    \vspace{-0.5cm}
\end{figure}

\begin{table*}[t]
\centering

\begin{minipage}[t]{0.47\textwidth}
\centering
\caption{AgileX Cobot Magic Platform Object Transport results over 10 trials per method.}
\label{tab:cobot-SingleObjMatch}

\vspace{2pt}

\small
\setlength{\tabcolsep}{4.0pt}
\renewcommand{\arraystretch}{1.12}

\begin{tabular}{@{}lcccc@{}}
\toprule
Method & Pick & Move \& Place & Right Container \\
\midrule
$\pi_{0.5}$
& 10/10 & 10/10 & 2/10  \\
Ours
& 10/10 & 9/10 & 8/10 \\

\bottomrule
\end{tabular}
\end{minipage}
\hfill
\begin{minipage}[t]{0.47\textwidth}
\centering
\caption{Simulated SingleMilkMatch ablation. Stage rates are cumulative.}
\label{tab:SingleMilkMatch-ablation}

\vspace{2pt}

\footnotesize
\setlength{\tabcolsep}{3.2pt}
\renewcommand{\arraystretch}{1.12}

\begin{tabular}{@{}lccc@{}}
\toprule
Variant & Pick & Move \& Place & Right Container\\
\midrule

$\pi_{0.5}$
& 0.70
& 0.60
& 0.20
\\

+ Planner
& 0.66
& 0.50
& 0.34
\\

+ VGGT-$\Omega$
& 0.70
& 0.64
& 0.40
\\
\hline
Ours
& 0.74
& 0.56
& 0.46
\\

\bottomrule
\end{tabular}
\end{minipage}
\end{table*}

\subsection{Ablation Study}
\label{sec:ablation}

We conduct an ablation study on the simulated SingleMilkMatch task to examine three components: planner-generated subtasks, VGGT-$\Omega$ conditioning, and learnable query tokens. Starting from $\pi_{0.5}$, we progressively add the high-level planner, direct VGGT-$\Omega$ modulation, and finally task-conditioned learnable queries with cross-attention. Table~\ref{tab:SingleMilkMatch-ablation} reports the resulting stage-level performance.
Terminal success increases from 0.20 for $\pi_{0.5}$ to 0.34 with planner-generated subtasks, showing that high-level memory substantially improves target selection. Adding VGGT-$\Omega$ conditioning further increases success to 0.40 and primarily improves low-level execution. The full model reaches 0.46 terminal success; among episodes in which the object is successfully placed, correct target selection rises to 0.82. These results suggest complementary roles for the two levels of the policy: structured memory supports history-dependent decision making, while task-conditioned geometric features improve execution and target-specific control.

\section{Conclusion and Limitations}
\label{sec:conclusion}
We introduced \benchmark, comprising nine mobile-manipulation tasks and 1,800 successful demonstrations organized around Environment Memory and Process Memory. Behavior-Conditioned Evaluation checks whether subsequent behavior satisfies constraints established by actual interactions, accommodating multiple valid execution paths. Our HMAP baseline combines structured semantic memory for planning with current multi-view geometry for control and improves average task success from 16.2\% to 24.2\% over $\pi_{0.5}$. These gains are concentrated in hidden-attribute recall, while improvements on process-memory tasks remain inconsistent and neither interleaved scheduling task is completed by the evaluated policies. These findings motivate further work on reliable memory updates, subtask switching, and resumption of pending actions.

Our evaluation is primarily conducted in simulation, with only one physical task evaluated over 10 trials per method, limiting conclusions about broader real-world performance. The five memory categories are not exhaustive, and the dataset contains only successful expert demonstrations. Stage-level metrics help localize failures but do not fully separate memory errors from execution errors. In SugarCount, stage applicability also varies across episodes, so SCR summarizes task progress over different episode subsets. More diagnostic evaluation protocols are therefore needed to distinguish history-dependent decision errors from physical execution failures.

\clearpage

\bibliography{references}
\bibliographystyle{references}

\clearpage
\appendix

\section{Task Specifications}
\label{app:tasks}

\begin{figure*}[t!]
    \centering
    \includegraphics[width=\textwidth,height=0.84\textheight,keepaspectratio]{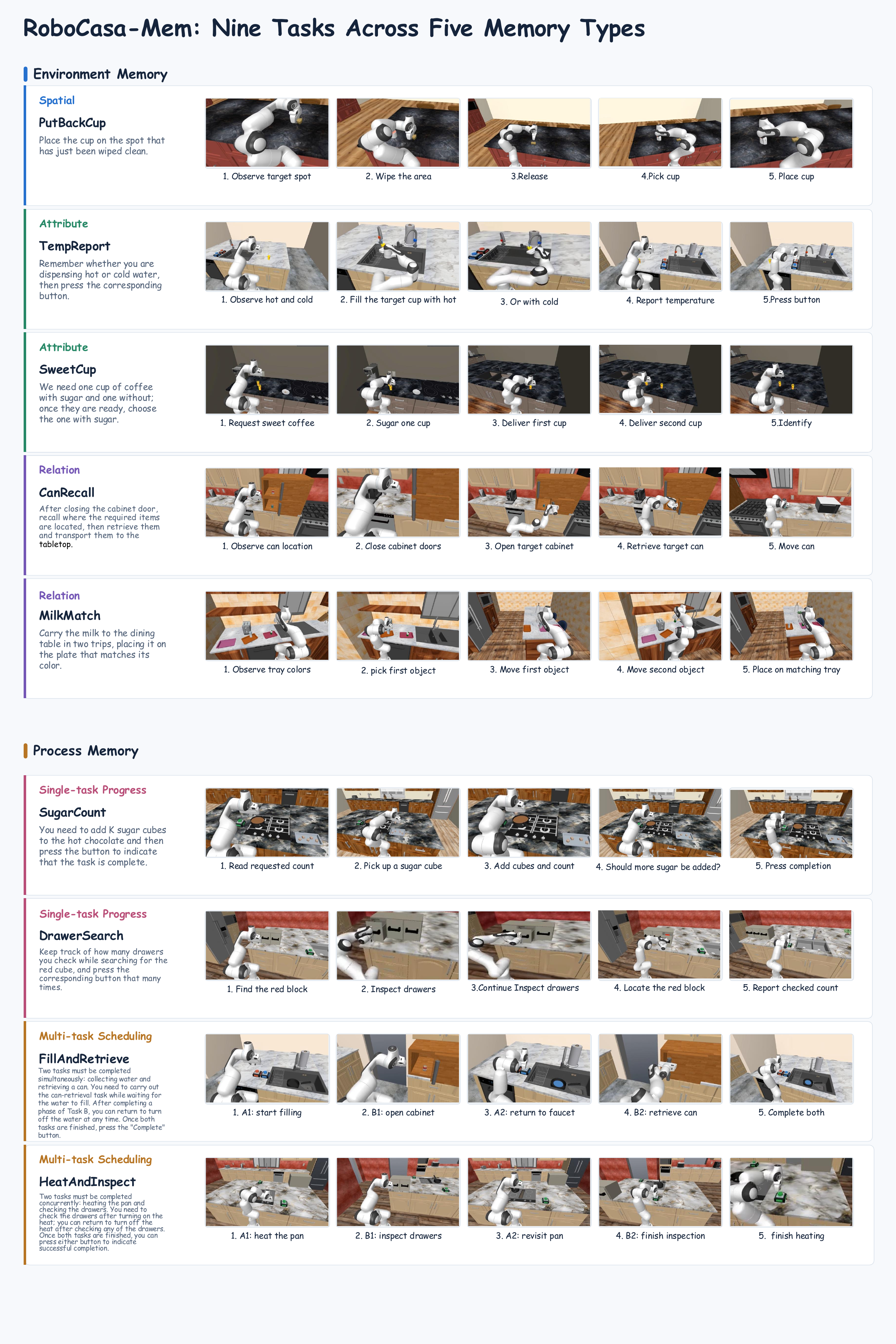}
    \caption{Execution workflows of the nine \benchmark tasks.}
    \label{fig:all-task}
\end{figure*}

The nine core tasks are summarized in Table~\ref{tab:tasks},
and their workflows are illustrated in Figure~\ref{fig:all-task}.
This section describes the information acquired during execution,
why it may become unavailable at a later decision point, and
what the reported evaluation stages assess.
These descriptions summarize task behavior rather than reproduce
the exact policy instructions.
A stage is credited only after its required predecessors have
been satisfied.
Final success additionally requires the task-specific terminal
semantic condition.

\subsection{Environment Memory Tasks}

\paragraph{PutBackCup (spatial memory).}
The robot observes the cup's initial location before moving it
aside to clean the area it occupied.
The cup's current position no longer identifies its original
location, which the robot must recall when returning it.
The reported stages check, in order, stain removal, placement
of the cleaning tool, and return of the cup to its original location.

\paragraph{SweetCup (attribute memory).}
The robot prepares one sweetened and one unsweetened cup.
It may choose which cup receives sugar, and the evaluator records
this choice from the actual rollout rather than assigning a
cup identity in advance.
After the sugar dissolves, the current RGB observation does not
reliably identify the sweetened cup.
The reported stages check sugar addition, staging of both cups,
and selection of the cup that received sugar.

\paragraph{TempReport (attribute memory).}
The robot dispenses hot or cold water and later reports which
temperature was dispensed.
At the reporting stage, the filled cup's appearance does not
directly reveal this attribute.
The stages check filling the cup, placing it in the service area,
and final task success, including the correct temperature report.

\paragraph{CanRecall (relational memory).}
The robot observes the associations between the requested cans
and their containing cabinets.
After closing both cabinet doors to hide their contents, it must
reopen the appropriate cabinet, retrieve the requested items,
and place them back on the tabletop.
The four reported stages check closure of either cabinet door
(Close-1), simultaneous closure of both doors (Close-2),
reopening of the correct cabinet after the contents have been
hidden (Reopen), and retrieval and placement of the requested
items back on the tabletop (Success).
Correct reopening assesses part of the recall decision,
while terminal success also depends on subsequent execution.

\paragraph{MilkMatch (relational memory).}
Two milk boxes are transported from colored source trays on one
side of the coffee machine to matching-color target trays on
the other side.
Turning toward the targets removes the source trays from the
immediate view.
Either box may be transported first.
The evaluator assigns ``first'' and ``second'' according to
the policy's actual pickup order and checks pickup, placement,
and correct color-matched placement for each box.
Terminal success requires both boxes to reach their corresponding
target trays.

\subsection{Process Memory Tasks}

\paragraph{DrawerSearch (single-task progress).}
The robot searches drawers for a red cube while tracking the
number of drawers inspected.
Finding the target determines when to stop searching, and the
inspection count determines the required number of button presses.
The stages check opening and closing a drawer, finding the target,
stopping the search after discovery, pressing the answer button,
and reporting the correct count.
The answer must reflect the drawers inspected during the
actual rollout.

\paragraph{SugarCount (single-task progress).}
The instruction specifies an episode-dependent target count $K$.
The robot tracks successful sugar-cube additions, stops at
the requested count, and presses the completion button.
The reported stages assess the first, second, and third additions,
followed by final task success.
Each addition stage applies only to episodes requesting at least
that many cubes, so its denominator may differ from those of
other addition stages.

\paragraph{HeatAndInspect (multi-task scheduling).}
The robot turns on the stove to heat a pan while inspecting drawers.
During inspection, it must retain the obligation to return
and turn off the stove before submission.
The stages check turning on the stove, inspecting a drawer,
completing either task thread, completing both threads,
and final task success.

\paragraph{FillAndRetrieve (multi-task scheduling).}
The robot starts dispensing water and retrieves a can while
the faucet is running.
Completion requires returning to turn off the faucet and
finishing the retrieval task.
The stages check turning on the faucet, retrieving the can,
completing either task thread, completing both threads,
and final task success.

\section{Data and Evaluation Details}
\label{app:data}

\paragraph{Observation and expert interfaces.}
Simulation trajectories are recorded at 20 Hz.
Each timestep contains three synchronized $224\times224$ RGB
images, a 37-dimensional robot state, and a 13-dimensional action.
The images are captured by the left and right external cameras
and the wrist-mounted camera.
The standard policy receives these RGB images, the robot state,
and the language instruction.
A rule-based expert combines MPLib for arm motion planning
with closed-loop mobile-base navigation.
Episode configurations vary task-relevant object positions
and appearances, navigation targets, manipulation order and
poses where applicable, and parameters such as the requested
sugar count.
The expert's internal task state is not included in the standard
policy inputs.

\paragraph{Storage and validation.}
Native HDF5 recordings store trajectories, camera recordings,
calibration information, and episode metadata.
Conversion to LeRobot preserves the alignment between each
observation and its subsequent action.
Termination and success flags are retained for bookkeeping
and evaluation but are excluded from the standard policy inputs.
We retain 200 successful demonstrations per core task.
Validators check the relevant grasp and release, containment,
fixture-state, count, and ordering conditions.
Trajectories that fail validation are excluded.
An action-only replay audit samples 20 accepted demonstrations
per task, yielding 180 sampled trajectories in total.

\paragraph{Control schema.}
Each 13-dimensional action comprises seven arm joint targets,
one gripper command, three mobile-base commands, one torso command,
and one base/arm mode indicator.
The exact axis order, units, mode encoding, and definitions
of the 37 robot-state dimensions should accompany the
released dataset.

\paragraph{Evaluation and aggregation.}
Methods are evaluated on 50 held-out scene seeds per core task,
using the same randomized environments across methods.
For task $j$, let $N_j$ denote the number of valid evaluation
episodes and $z_{ij}$ indicate terminal semantic success
in episode $i$.
Let $K_j$ denote the number of evaluation stages.
For stage $k$, let $A_{jk}$ be the applicable episode set and
$s_{ijk}$ indicate whether the stage predicate and its required
predecessors have been satisfied.
We compute
\begin{equation}
    \mathrm{SR}_j
    = \frac{1}{N_j}\sum_{i=1}^{N_j}z_{ij},
    \qquad
    r_{jk}
    = \frac{\sum_{i\in A_{jk}}s_{ijk}}{|A_{jk}|},
    \qquad
    \mathrm{SCR}_j
    = \frac{1}{K_j}\sum_{k=1}^{K_j}r_{jk}.
\end{equation}
An episode that fails before reaching an applicable stage
remains in that stage's denominator.
SugarCount addition stages have different applicable episode
sets when target counts differ, so their rates need not
decrease monotonically.
The overall SR and SCR in Table~\ref{tab:main-results} are
the unweighted means of the nine per-task values.

When adjacent stages share the same applicable episode set,
have a strict predecessor relation, and satisfy $r_{j,k-1}>0$,
the ratio $r_{jk}/r_{j,k-1}$ measures completion of stage $k$
among episodes that completed stage $k-1$.
For $\pi_{0.5}$ and HMAP, respectively, the conditional rates
are 0.72 and 0.83 for returning the cup in PutBackCup,
0.22 and 0.45 for selecting the sweetened cup in SweetCup,
and 0.41 and 0.90 for the final reporting stage in TempReport.
For MilkMatch, the conditional rates of correct placement
given placement are 0.48 and 0.77 for the first box and
0.46 and 0.80 for the second.
A predicate that combines target selection with manipulation
cannot attribute every failure uniquely to memory.

\section{Method and Annotation Details}
\label{app:method}

\paragraph{Planner state.}
The Qwen3-VL-4B-Instruct planner~\citep{bai2025qwen3vl}
receives the instruction, previous entity and process memories,
and observations at times $t_k$ and $t_k-50$, as defined in
Eq.~\eqref{eq:entity-memory-update}.
Entity memory records relevant properties, locations, and
relations, such as which cup received sugar or a milk box's
source tray.
Process memory records completed steps and pending obligations,
such as inspected drawers or a faucet that remains to be
turned off.
Each planner call returns updated memories and one executable
language subgoal.
The memories are passed to the next planner call, while only
the subgoal is supplied to the low-level controller.
Structured supervision is derived offline from demonstration
trajectories and checked for consistency with task-event ordering.

\paragraph{Geometric conditioning.}
The low-level controller builds on
$\pi_{0.5}$~\citep{black2025pi05}.
A frozen VGGT-$\Omega$ encoder~\citep{wang2026vggtomega}
produces multi-view geometric tokens separately from the
standard visual--language--state representation.
Eight learnable query embeddings are processed together with
the current task context.
The resulting representations attend to the geometric tokens
through cross-attention, and the fused features provide scale
and shift modulation for the action expert.
The planner and controller are trained separately, with the
controller optimized using the standard flow-matching objective
for action generation.

\paragraph{Reported training schedules.}
The reported batch size is 48.
ACT and DP are trained for 40{,}000 optimization steps
with a sampling stride of 10.
The reported training budgets for $\pi_{0.5}$ and HMAP are
20{,}000 steps for PutBackCup, TempReport, DrawerSearch,
and SugarCount, and 30{,}000 steps for SweetCup, CanRecall,
MilkMatch, and HeatAndInspect.
The available schedule does not specify the training budget
for FillAndRetrieve.
The reported replanning interval is 40 control steps for
MilkMatch, DrawerSearch, SugarCount, and PutBackCup.
Optimizer settings, LoRA configurations, checkpoint-selection
criteria, and the remaining task-specific settings remain
to be specified from the final run configurations.

\clearpage
\section{Complete Results}
\suppressfloats[t]
\label{app:results}

Tables~\ref{tab:stage-results-part1}
and~\ref{tab:stage-results-part2} report stage-level completion
rates for all nine benchmark tasks, with stage descriptions
provided below each subtable.
Final task success rates are summarized separately in
Table~\ref{tab:main-results}.

\begin{table*}[!htbp]
\centering
\caption{Stage-level completion rates for PutBackCup, SweetCup,
and TempReport. Each value is the fraction of evaluated episodes
satisfying the corresponding stage and its required predecessors.}
\label{tab:stage-results-part1}

\vspace{2pt}

\begin{minipage}[t]{0.31\textwidth}
\centering

\textbf{(a) PutBackCup}

\vspace{2pt}

\scriptsize
\setlength{\tabcolsep}{3.5pt}
\renewcommand{\arraystretch}{1.12}

\begin{tabular}{@{}lccc@{}}
\toprule
Method & Clean & Tool & Return \\
\midrule
ACT         & 0.72 & 0.72 & 0.00 \\
DP          & 0.30 & 0.30 & 0.00 \\
$\pi_{0.5}$ & 0.94 & 0.94 & 0.68 \\
MemER-style & 0.28 & 0.28 & 0.10 \\
HMAP (Ours) & 0.84 & 0.84 & 0.70 \\
\bottomrule
\end{tabular}

\vspace{2pt}

\raggedright
\scriptsize
\textbf{Stages.} Clean the stain, put down the cleaning tool,
and return the cup.
\end{minipage}
\hfill
\begin{minipage}[t]{0.36\textwidth}
\centering

\textbf{(b) SweetCup}

\vspace{2pt}

\scriptsize
\setlength{\tabcolsep}{3.0pt}
\renewcommand{\arraystretch}{1.12}

\begin{tabular}{@{}lcccc@{}}
\toprule
Method & Sugar & Stage-1 & Stage-2 & Select \\
\midrule
ACT         & 0.22 & 0.04 & 0.02 & 0.00 \\
DP          & 0.10 & 0.04 & 0.00 & 0.00 \\
$\pi_{0.5}$ & 0.48 & 0.38 & 0.18 & 0.04 \\
MemER-style & 0.72 & 0.12 & 0.00 & 0.00 \\
HMAP (Ours) & 0.82 & 0.68 & 0.22 & 0.10 \\
\bottomrule
\end{tabular}

\vspace{2pt}

\raggedright
\scriptsize
\textbf{Stages.} Add sugar, stage the first cup,
stage the second cup, and select the sweetened cup.
\end{minipage}
\hfill
\begin{minipage}[t]{0.31\textwidth}
\centering

\textbf{(c) TempReport}

\vspace{2pt}

\scriptsize
\setlength{\tabcolsep}{3.5pt}
\renewcommand{\arraystretch}{1.12}

\begin{tabular}{@{}lccc@{}}
\toprule
Method & Fill & Place & Success \\
\midrule
ACT         & 0.26 & 0.00 & 0.00 \\
DP          & 0.38 & 0.18 & 0.08 \\
$\pi_{0.5}$ & 0.94 & 0.54 & 0.22 \\
MemER-style & 0.98 & 0.76 & 0.00 \\
HMAP (Ours) & 0.98 & 0.96 & 0.86 \\
\bottomrule
\end{tabular}

\vspace{2pt}

\raggedright
\scriptsize
\textbf{Stages.} Fill the cup, place it in the service area,
and achieve final task success.
\end{minipage}

\end{table*}

\begin{table*}[!htbp]
\centering
\caption{Stage-level completion rates for the remaining six
\benchmark tasks. Abbreviated stage names are explained below
each subtable.}
\label{tab:stage-results-part2}

\vspace{2pt}

\begin{minipage}[t]{0.47\textwidth}
\centering

\textbf{(d) CanRecall}

\vspace{2pt}

\fontsize{8}{9}\selectfont
\setlength{\tabcolsep}{2.5pt}
\renewcommand{\arraystretch}{1.12}

\begin{tabular}{@{}lcccc@{}}
\toprule
Method & Close-1 & Close-2 & Reopen & Success \\
\midrule
ACT         & 0.34 & 0.00 & 0.00 & 0.00 \\
DP          & 0.18 & 0.00 & 0.00 & 0.00 \\
$\pi_{0.5}$ & 0.76 & 0.14 & 0.04 & 0.00 \\
MemER-style & 0.80 & 0.02 & 0.00 & 0.00 \\
HMAP (Ours) & 0.80 & 0.24 & 0.00 & 0.00 \\
\bottomrule
\end{tabular}

\vspace{2pt}

\raggedright
\scriptsize
\textbf{Stages.} Close either cabinet door (Close-1), have both
doors closed simultaneously (Close-2), reopen the correct cabinet
after hiding its contents (Reopen), and retrieve the requested
items and place them back on the tabletop (Success).
\end{minipage}
\hfill
\begin{minipage}[t]{0.47\textwidth}
\centering

\textbf{(e) MilkMatch}

\vspace{2pt}

\fontsize{8}{9}\selectfont
\setlength{\tabcolsep}{2.5pt}
\renewcommand{\arraystretch}{1.12}

\begin{tabular}{@{}lcccccc@{}}
\toprule
Method & P1 & D1 & C1 & P2 & D2 & C2 \\
\midrule
ACT         & 0.10 & 0.10 & 0.00 & 0.00 & 0.00 & 0.00 \\
DP          & 0.20 & 0.12 & 0.00 & 0.00 & 0.00 & 0.00 \\
$\pi_{0.5}$ & 0.92 & 0.80 & 0.38 & 0.34 & 0.26 & 0.12 \\
MemER-style & 0.52 & 0.42 & 0.14 & 0.02 & 0.02 & 0.00 \\
HMAP (Ours) & 0.84 & 0.70 & 0.54 & 0.36 & 0.20 & 0.16 \\
\bottomrule
\end{tabular}

\vspace{2pt}

\raggedright
\scriptsize
\textbf{Stages.} P, D, and C denote pickup, placement,
and correct placement, respectively.
The indices 1 and 2 refer to execution order.
\end{minipage}

\vspace{7pt}

\begin{minipage}[t]{0.47\textwidth}
\centering

\textbf{(f) DrawerSearch}

\vspace{2pt}

\fontsize{8}{9}\selectfont
\setlength{\tabcolsep}{2.5pt}
\renewcommand{\arraystretch}{1.12}

\begin{tabular}{@{}lccccc@{}}
\toprule
Method & Open/Close & Find & Stop & Button & Count \\
\midrule
ACT         & 0.34 & 0.04 & 0.04 & 0.00 & 0.00 \\
DP          & 0.22 & 0.10 & 0.06 & 0.00 & 0.00 \\
$\pi_{0.5}$ & 0.84 & 0.38 & 0.16 & 0.12 & 0.12 \\
MemER-style & 0.86 & 0.38 & 0.02 & 0.00 & 0.00 \\
HMAP (Ours) & 0.84 & 0.38 & 0.16 & 0.08 & 0.08 \\
\bottomrule
\end{tabular}

\vspace{2pt}

\raggedright
\scriptsize
\textbf{Stages.} Open and close a drawer, find the target,
stop searching, press the button, and press it the correct
number of times.
\end{minipage}
\hfill
\begin{minipage}[t]{0.47\textwidth}
\centering

\textbf{(g) SugarCount}

\vspace{2pt}

\fontsize{8}{9}\selectfont
\setlength{\tabcolsep}{2.5pt}
\renewcommand{\arraystretch}{1.12}

\begin{tabular}{@{}lcccc@{}}
\toprule
Method & Cube-1 & Cube-2 & Cube-3 & Success \\
\midrule
ACT         & 0.22 & 0.00  & 0.00  & 0.04 \\
DP          & 0.50 & 0.125 & 0.00  & 0.14 \\
$\pi_{0.5}$ & 0.46 & 0.04  & 0.125 & 0.28 \\
MemER-style & 0.26 & 0.04  & 0.00  & 0.08 \\
HMAP (Ours) & 0.48 & 0.04  & 0.00  & 0.28 \\
\bottomrule
\end{tabular}

\vspace{2pt}

\raggedright
\scriptsize
\textbf{Stages.} Add the first, second, and third sugar cubes,
followed by final task success.
Sugar-addition stages may apply to different episode subsets.
\end{minipage}

\vspace{7pt}

\begin{minipage}[t]{0.47\textwidth}
\centering

\textbf{(h) HeatAndInspect}

\vspace{2pt}

\fontsize{8}{9}\selectfont
\setlength{\tabcolsep}{2.5pt}
\renewcommand{\arraystretch}{1.12}

\begin{tabular}{@{}lccccc@{}}
\toprule
Method & Stove & Inspect & Either & Both & Success \\
\midrule
ACT         & 1.00 & 0.24 & 0.00 & 0.00 & 0.00 \\
DP          & 0.98 & 0.18 & 0.00 & 0.00 & 0.00 \\
$\pi_{0.5}$ & 0.98 & 0.54 & 0.08 & 0.00 & 0.00 \\
MemER-style & 1.00 & 0.58 & 0.04 & 0.00 & 0.00 \\
HMAP (Ours) & 1.00 & 0.88 & 0.62 & 0.00 & 0.00 \\
\bottomrule
\end{tabular}

\vspace{2pt}

\raggedright
\scriptsize
\textbf{Stages.} Turn on the target stove, inspect a drawer,
complete either task thread, complete both threads,
and achieve final task success.
\end{minipage}
\hfill
\begin{minipage}[t]{0.47\textwidth}
\centering

\textbf{(i) FillAndRetrieve}

\vspace{2pt}

\fontsize{8}{9}\selectfont
\setlength{\tabcolsep}{2.5pt}
\renewcommand{\arraystretch}{1.12}

\begin{tabular}{@{}lccccc@{}}
\toprule
Method & Faucet & Retrieve & Either & Both & Success \\
\midrule
ACT         & 1.00 & 0.18 & 0.08 & 0.02 & 0.00 \\
DP          & 0.90 & 0.00 & 0.00 & 0.00 & 0.00 \\
$\pi_{0.5}$ & 1.00 & 0.46 & 0.40 & 0.08 & 0.00 \\
MemER-style & 1.00 & 0.00 & 0.00 & 0.00 & 0.00 \\
HMAP (Ours) & 1.00 & 0.40 & 0.32 & 0.00 & 0.00 \\
\bottomrule
\end{tabular}

\vspace{2pt}

\raggedright
\scriptsize
\textbf{Stages.} Turn on the faucet, retrieve the can,
complete either task thread, complete both threads,
and achieve final task success.
\end{minipage}

\end{table*}

\end{document}